\documentclass[10pt,journal,compsoc]{IEEEtran}

\usepackage{amsmath}
\usepackage{graphicx}
\usepackage{amssymb}
\usepackage{booktabs}
\usepackage{algorithmic}
\usepackage[ruled]{algorithm2e}
\usepackage{soul}
\usepackage{color}
\usepackage{float}
\usepackage{wrapfig}
\usepackage{url}
\usepackage{array}
\usepackage{bm}
\usepackage{multirow}
\usepackage{tabularx,booktabs}
\usepackage{xcolor,colortbl}

\definecolor{myorange}{rgb}{0.98,0.86,0.58}
\definecolor{myyellow}{rgb}{0.99,0.99,0.82}
\newcommand{\cello}[1]{\cellcolor{myorange} #1}
\newcommand{\celly}[1]{\cellcolor{myyellow} #1}

\ifCLASSOPTIONcompsoc
  \usepackage[nocompress]{cite}
\else
  \usepackage{cite}
\fi

\ifCLASSINFOpdf
\else
\fi

\begin{document}

\title{Interactive Rendering of Relightable and Animatable Gaussian Avatars}

\author{Youyi~Zhan,
        Tianjia~Shao,
        He~Wang,
        Yin~Yang,
        and Kun~Zhou,~\IEEEmembership{Fellow,~IEEE}
\IEEEcompsocitemizethanks{\IEEEcompsocthanksitem Youyi Zhan, Tianjia Shao, Kun Zhou are with the State Key Lab of CAD \& CG, Zhejiang University, Hangzhou 310058, China. Tianjia Shao is the corresponding author of the work. \protect\\ E-mail: \{zhanyy, tjshao\}@zju.edu.cn, kunzhou@acm.org. 
\IEEEcompsocthanksitem He Wang is with UCL Centre for Artificial Intelligence, Department of Computer Science, University College London, Gower Street London, WC1E 6BT United Kingdom. \protect\\ E-mail: he\_wang@ucl.ac.uk. 
\IEEEcompsocthanksitem Yin Yang is with Kahlert School of Computing, University of Utah, USA. E-mail: yangzzzy@gmail.com.}
\thanks{Manuscript received April 19, 2005; revised August 26, 2015.}}

\markboth{IEEE TRANSACTIONS ON VISUALIZATION AND COMPUTER GRAPHICS}%
{Shell \MakeLowercase{\textit{et al.}}: Bare Demo of IEEEtran.cls for Computer Society Journals}

\IEEEtitleabstractindextext{%
\begin{abstract}

Creating relightable and animatable avatars from multi-view or monocular videos is a challenging task for digital human creation and virtual reality applications. Previous methods rely on neural radiance fields or ray tracing, resulting in slow training and rendering processes. By utilizing Gaussian Splatting, we propose a simple and efficient method to decouple body materials and lighting from sparse-view or monocular avatar videos, so that the avatar can be rendered simultaneously under novel viewpoints, poses, and lightings at interactive frame rates (6.9 fps). Specifically, we first obtain the canonical body mesh using a signed distance function and assign attributes to each mesh vertex. The Gaussians in the canonical space then interpolate from nearby body mesh vertices to obtain the attributes. We subsequently deform the Gaussians to the posed space using forward skinning, and combine the learnable environment light with the Gaussian attributes for shading computation. To achieve fast shadow modeling, we rasterize the posed body mesh from dense viewpoints to obtain the visibility. Our approach is not only simple but also fast enough to allow interactive rendering of avatar animation under environmental light changes. Experiments demonstrate that, compared to previous works, our method can render higher quality results at a faster speed on both synthetic and real datasets. 

\end{abstract}

}

\maketitle

\IEEEdisplaynontitleabstractindextext

\IEEEpeerreviewmaketitle



\IEEEraisesectionheading{\section{Introduction}\label{sec:introduction}}

\IEEEPARstart{C}{reating} realistic human avatars is a challenging problem and widely used in various fields, such as virtual reality and visual content creation. To achieve high visual realism, the avatar should be able to be animated under various poses and lightings. Existing methods of creating the relightable avatar involve capturing dense view videos in a light stage with controllable illuminations (OLAT light) and decoupling the materials and the environment light~\cite{guo2019relightables,zhang2021neural,yang2023towards,sarkar2023litnerf,bi2021deep,saito2023relightable}. However, the expensive devices and settings are not accessible easily, restricting the application of these methods. By learning from multi-view RGB videos, many works have successfully modeled high quality digital avatars using neural radiance fields (NeRF~\cite{mildenhall2021nerf}) or 3D Gaussian Spaltting (3DGS~\cite{kerbl20233d}), but these works fail to generalize to unseen lighting conditions. This key limitation is due to that they bake the view-dependent color onto the Gaussian or neural field without considering the intrinsic material properties. 

Recent works attempt to decouple body's materials from the videos captured under unknown illumination, thus enabling relighting under novel environmental light.
These methods are usually based on neural volume rendering, which defines a neural human in canonical space and obtains the material properties by inferring from the neural network. Shading is computed by casting rays from the camera, sampling space points, and inversely wrapping them to canonical space to obtain the material properties, which are then evaluated by the rendering equation. Although this neural rendering technique has already achieved good rendering effects on relightable humans, the design is inherently slow in both training and rendering. This is because the pipeline uses multilayer perceptron (MLP) networks to encode the scene information, which need to be inferred many times to obtain the density and color for further volume rendering, to the extent that it would take a considerable amount of time. Even though some methods have adopted feature encoding based on iNGP~\cite{muller2022instant} to accelerate the inference process, rendering an image still requires sampling each pixel many times, making the time of obtaining a rendered result excessively long. The efficiency problem becomes worsened when the shadow effect is involved, as these works use explicit ray tracing~\cite{bolanos2024gaussian,wang2023intrinsicavatar,li2024animatable}, soft shadows~\cite{xu2023relightable} or pretrained models~\cite{chen2022relighting4d,lin2024relightable} to calculate the visibility, further slowing down the speed.

In this paper, we propose to create a relightable and animatable avatar from multi-view or monocular videos, which can render high-quality avatar animation under environmental light changes at interactive frame rates (6.9 fps). 3DGS has succeeded in modeling high-quality animatable avatars at real-time frame rates~\cite{moreau2023human,hu2023gauhuman,qian20233dgs,wen2024gomavatar,kocabas2023hugs,pang2023ash,li2023animatable,jiang2024uv,hu2023gaussianavatar,shao2024splattingavatar,wen2024gomavatar}, so we adopt it. However, it's not trivial to incorporate 3DGS to build the relightable avatar efficiently. Currently, it faces the following challenges. First, in our relighting task, we need to encode information for each Gaussian such that shading color can be calculated under different lighting conditions. Second, we need to solve the problem of how to render the relightable human efficiently, especially when in the presence of the shadows caused by body self-occlusion.

To address the aforementioned issues, we propose a new Gaussian representation for relighting avatars during animation. We first define a body mesh and Gaussians in the canonical space, as shown in Figure~\ref{fig:pipeline}, and the mesh vertices and Gaussians can be animated to the posed space via a body model (i.e., SMPL~\cite{loper2015smpl}). 

The Gaussian properties are interpolated from the mesh vertex properties, which include basic Gaussian properties (position, rotation, scale, and opacity) and material properties.
During Gaussian optimization, we optimize the basic Gaussian properties, the material properties and the environment light simultaneously so that the new shading color could be computed under novel illuminations. 

Specifically, we initially train a signed distance function (SDF) to obtain the canonical body mesh and initialize the Gaussian primitives near the body mesh. The body mesh vertices contain attributes, including basic Gaussian properties and material properties (albedo, roughness, specular tint and visibility), LBS weight, normal and position displacement. For each Gaussian, its attributes are interpolated from those of nearby mesh vertices. During animation, Gaussians are first added with position displacements and then deformed to the posed space via forward LBS. In the posed space, the shading color of the Gaussian is computed by explicitly integrating the rendering equation, and then fed into the Gaussian renderer to output the final image. The entire process is differentiable, allowing us to optimize the material properties and environment map directly by gradient-based optimization. During the training process, we propose an additional densification method to control the Gaussian density over the mesh surface, preventing holes in novel view synthesis (see Figure~\ref{fig:ablation_compare}), and add a scale loss to avoid artifacts caused by the stretched Gaussians (see Figure~\ref{fig:ablation_scale}). For visibility computation, we deform the body mesh to the posed space and rasterize the mesh from dense view directions to calculate the visibility of the mesh vertices at a given pose and view direction. As the hardware-accelerated rasterization is utilized for rapid visibility calculation, it allows the relighting computation at interactive frame rates. Further, we achieve single-view editing of human appearance by optimizing attributes, allowing users to easily customize the human appearance.

We evaluate our approach both quantitatively and qualitatively using synthetic and real datasets. Compared to previous state-of-the-art works, our method can provide better rendering quality in novel pose synthesis and under novel illuminations. Ablation studies have demonstrated the effectiveness of our design in enhancing relighting results. We also demonstrate that the rendering speed is fast enough to visualize the relighting results interactively.



\section{Related Work}

\subsection{Human Avatar}

Creating digital humans from real world data is in high demand but challenging. Previous methods often use complex capturing devices, such as dense camera arrays~\cite{collet2015high,xiang2022dressing,xiang2021modeling} or depth cameras~\cite{tong2012scanning,bogo2015detailed,habermann2019livecap,xiang2023drivable}, to obtain high-quality human, enabling free-viewpoint rendering. However, not everyone has access to such devices, and some works~\cite{tong2012scanning,bogo2015detailed,collet2015high} cannot generate animatable digital humans, limiting their usage. In recent years, many works~\cite{peng2021neural,weng2022humannerf,wang2022arah,jiang2022neuman,zheng2022structured,yu2023monohuman,peng2021animatable,yang2022banmo,zheng2022structured,jiang2023instantavatar,guo2023vid2avatar} have used NeRF~\cite{mildenhall2021nerf} to represent the human body by learning from multi-view videos, achieving pleasant rendering results. They usually define the human body as articulated using the SMPL body model~\cite{loper2015smpl} in the neural canonical space and warp the observed position to the canonical space based on inverse LBS to obtain the attributes (like color and density) for further volume rendering. The reconstructed neural body can be rendered in a novel view and can also be driven by new poses. However, neural volume rendering requires multiple samplings for each pixel and conducting inverse LBS many times, which is time-consuming to render a single image. Even if \cite{jiang2023instantavatar} adopts iNGP~\cite{muller2022instant} for fast training and rendering, it's still difficult to simultaneously achieve high resolution, high quality, and high speed.

3DGS~\cite{kerbl20233d} has achieved outstanding results in scene reconstruction and high-quality rendering from novel views. Its explicit Gaussian representation is highly efficient and has been widely used in multi-view human body reconstruction and rendering as well. Similar to NeRF-based methods, the 3DGS-based works define the Gaussians in the canonical space and use MLP~\cite{moreau2023human,hu2023gauhuman,qian20233dgs,wen2024gomavatar}, feature grid~\cite{kocabas2023hugs}, convolutional neural network (CNN)~\cite{pang2023ash,li2023animatable,jiang2024uv,hu2023gaussianavatar} or mesh~\cite{shao2024splattingavatar,wen2024gomavatar} to decode the Gaussian properties. Then, the Gaussians are deformed to the posed space by forward LBS and rasterized for the final images. They all achieve high-quality and real-time rendering under novel views and poses. However, whether they are NeRF-based or Gaussian-based methods, they all model the view-dependent or pose-dependent color when reconstructing the human body, and do not separate the lighting from the human body's material. Thus, they cannot achieve relighting under novel lighting conditions.

\subsection{Radiance-field-based Inverse Rendering}

Inverse rendering aims to recover the material of an object from multi-view photos, enabling it to be relighted under novel lighting conditions. Many studies use NeRF to solve the inverse rendering problem~\cite{srinivasan2021nerv,zhang2021nerfactor,zhang2022modeling,hasselgren2022shape,jin2023tensoir,wu2023nerf,ling2024nerf}. They typically encode geometric and material information into the radiance field, learn the outgoing radiance from captured images, and optimize the materials to decouple the environmental light and materials. \cite{zhang2022modeling,hasselgren2022shape,wu2023nerf} use SDF to reconstruct and separate explicit geometric information for better material decoupling. Zhang et al.~\cite{zhang2022modeling} can model the indirect illumination. PhySG~\cite{zhang2021physg} represents specular BRDFs and environmental illumination using mixtures of spherical Gaussians. NeRD~\cite{boss2021nerd} and Neural-PIL~\cite{boss2021neural} can decouple materials using images captured under different illumination conditions. Lyu et al.~\cite{lyu2022neural} generate relighting results with global illumination. NeRO~\cite{liu2023nero} can reconstruct the BRDF of reflective objects with strong reflective appearances. Mai et al.~\cite{mai2023neural} use a microfacet reflectance model to recover high-quality materials, geometry, and illumination, while NeMF~\cite{zhang2023nemf} uses microflake volume to relight complex objects.

3DGS also propels the development of inverse rendering, achieving higher quality rendering for free view relighting. Existing works~\cite{shi2023gir,gao2023relightable,liang2023gs,wu2024deferredgs} assign the material attributes to Gaussians and optimize the attributes to decouple the lighting and materials. Unlike the neural-based method, which can obtain normal information through gradients of the neural field, obtaining the normal from a Gaussian scene is not straightforward. GIR~\cite{shi2023gir} observes that the maximum cross-section of a Gaussian contributes most to the rendered color, so it defines the shortest axis of the Gaussian as the normal direction. Gao et al.~\cite{gao2023relightable} outputs the depth and normal maps and constrains the two maps to be consistency. GS-IR~\cite{liang2023gs} estimates the scene depth to obtain a rough normal map, which is further optimized in subsequent steps. DeferredGS~\cite{wu2024deferredgs} uses SDF to obtain geometry and derives normals from SDF gradients. While all the aforementioned works achieve good results, they are only suitable for static objects and have not been specifically designed for dynamic human bodies.

\subsection{Human Relighting}

To reconstruct a relightable human, the key step is to restore the materials. By leveraging the priors learned from a large amount of data, some methods attempt to infer materials from a human photo and further perform image-based relighting on the faces~\cite{wang2020single,sun2019single,zhou2019deep,yeh2022learning,meka2020deep,mei2024holo}, upper body~\cite{pandey2021total,kim2024switchlight} or full body~\cite{yoshihiro2018relighting,ji2022geometry}. Due to the absence of underlying geometric information, such a design cannot alter the viewpoint and poses of the human body. On the other hand, some methods rely on a light stage with dense cameras and controllable lighting~\cite{guo2019relightables,zhang2021neural,yang2023towards,sarkar2023litnerf,bi2021deep,saito2023relightable}, which can restore high-quality human materials and achieve excellent relighting results. However, such an approach relies on extensive setups, which are costly and not publicly available. For this reason, an increasing number of studies are exploring how to reconstruct relightable human bodies under in-the-wild illuminations using sparse or monocular viewpoints. 

Relighting4D~\cite{chen2022relighting4d} proposes a method to reconstruct human geometry, recover human materials under unknown environmental lighting, and perform relighting. This method assigns latent features to each frame, which are then used in the MLP for modeling appearance and occlusion maps. Therefore, Relighting4D cannot transfer to novel poses. Sun et al.\cite{sun2023neural} uses inverse mapping to transform points from the observation space to the canonical space to obtain the material attributes, which are then used for shading color calculation. RANA~\cite{iqbal2023rana} fits the person via the SMPL+D model and trains networks to further refine albedo and normal. However, neither Sun et al. or RANA models the visibility, therefore the relighted body lacks shadows. To model shadows, many methods provide their unique solutions. RelightableAvatar-Lin~\cite{lin2024relightable} trains the part-wise MLP to estimate visibility under novel poses. Bolanos et al.~\cite{bolanos2024gaussian} uses a Gaussian density model as an approximation to the NeRF’s density field for easier visibility calculation. RelightableAvatar-Xu~\cite{xu2023relightable} proposes Hierarchical Distance Query on the SDF field for sphere tracing, and further utilizes Distance Field Soft Shadow (DFSS) for soft visibility. IntrinsicAvatar~\cite{wang2023intrinsicavatar} implements the body inverse rendering by explicit ray tracing and introducing the secondary shading effects, thus modeling accurate shadows naturally. The above methods successfully model the shadows generated by body occlusion, achieving realistic relighting results. However, these methods are based on neural fields, which require expensive ray marching with many sample point queries with MLPs, therefore slowing down the speed. They all require several seconds for an image, restricting their applications in non performance critical applications. RGCA~\cite{saito2023relightable} is able to create realistic and relightable head avatar, but this method requires to capture the data in a light stage with controllable lights and hundreds of cameras, which is difficult for ordinary users to obtain. By using 3DGS, Li et al.~\cite{li2024animatable} achieves excellent relighting results with unique body details under novel poses. However, it is trained under relatively dense viewpoints and relies on ray tracing to calculate visibility, which still requires several seconds to render a picture. MeshAvatar~\cite{chen2024meshavatar} uses a UNet to model the pose-dependent material fields, and Monte-Carlo sampling to compute the outgoing radiance. However, it cannot recover fine geometry under sparse-view settings, resulting in suboptimal relighting results (see Figure~\ref{fig:compare}).



\section{Method}

Given $N_c$ viewpoint avatar videos under unknown lighting with $T$ frames $ \{ I^{t,c} \}_{ t\in [1,T], c\in [1,N_c] }$ and avatar's body poses for each frame $ \{ \theta^t \}_{t \in [1,T]}$, our goal is to construct an animatable and relightable avatar $\mathcal{A}$ and render the relighting image $I'$ given a novel pose $\theta'$ and new environment map $L'$ at interactive frame rates. The avatar $\mathcal{A} = \{ M, G, L \}$ contains three components: a canonical body mesh with attributes on the vertices $M=\{ f^i, *^j \}_{i\in [1,N_f], j\in [1,N_v]}$, where $f$ is the mesh triangle and $*$ are the vertex attributes; a set of Gaussians floating near the body mesh, with attributes on Gaussians as well $G=\{ *_g^i \}_{i\in [1,N_g]}$, where $*_g$ are the Gaussian attributes; and a learnable environment lighting $L \in \mathbb{R}^{16\times 32 \times 3}$. The following table gives the notations used in the following sections. Our optimizing goal is to recover the above attributes and the unknown lighting from the given videos to achieve the relightable avatar.
\begin{table}[htbp]
\resizebox{\columnwidth}{!}{
\begin{tabular}{lp{15em}|lp{15em}}
\toprule
$\mathbf{x}$ / $\mathbf{x}_g$      &      Vertex/Gaussian position  &
$\mathbf{r}$ / $\mathbf{r}_g$      &      Vertex/Gaussian rotation  \\ 
$\mathbf{s}$ / $\mathbf{s}_g$      &      Vertex/Gaussian scale  &
$\mathbf{a}$ / $\mathbf{a}_g$      &      Vertex/Gaussian albedo  \\ 
$\gamma$ / $\mathbf{\gamma}_g$      &      Vertex/Gaussian roughness  & 
$p$ / $p_g$      &      Vertex/Gaussian specular tint  \\ 
$\mathbf{w}$ / $\mathbf{w}_g$      &      Vertex/Gaussian LBS weight  &
$\mathbf{n}$ / $\mathbf{n}_g$      &      Vertex/Gaussian normal  \\ 
$\Delta \mathbf{x}$ / $\Delta \mathbf{x}_g$      &      Vertex/Gaussian displacement  & 
$\mathbf{v}$ / $\mathbf{v}_g$      &      Vertex/Gaussian visibility  \\ 
$\bar{\mathbf{x}}_g$ & Gaussian position with displacement &
$\bar{\mathbf{x}}_{gd}$ & Gaussian position with displacement, at posed space \\
$\mathbf{n}_{gd}$ & Gaussian normal at posed space &
$\mathbf{r}_{gd}$ & Gaussian rotation at posed space \\
\bottomrule
\end{tabular}}
\end{table}

\begin{figure*}[t]
  \begin{center}
    \includegraphics[width=0.975\textwidth]{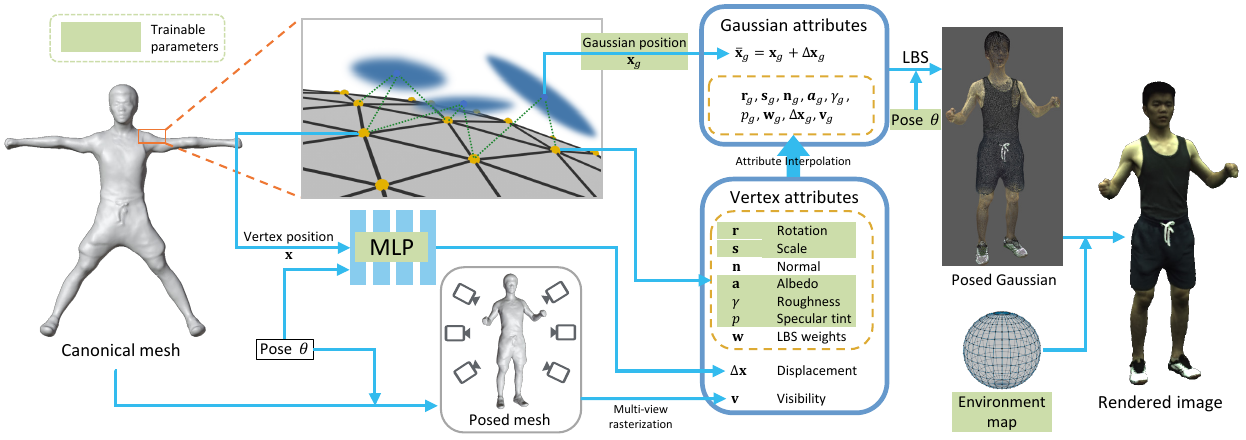}
  \end{center}
  \vspace{-5mm}
  \caption{Pipeline overview. Starting from the canonical mesh reconstructed from SDF, Gaussians are initialized near the mesh surface. The attributes of the Gaussians are interpolated from the neighboring vertices. Then Gaussians are deformed to the posed space and rasterized to produce an image (Section~\ref{sec:gaussianavatar}). Visibility is obtained from multi-view rendering of the posed mesh to model shadows (Section~\ref{sec:visibility}). Through photometric loss and other constraints, the environmental light and body materials can be separated for further relighting (Section~\ref{sec:training}).
  }
  \label{fig:pipeline}
\end{figure*}

\subsection{Relightable Gaussian Avatar}
\label{sec:gaussianavatar}

Our method starts with the reconstructed body mesh serving as the proxy for our expression. Previous methods have successfully constructed the body geometry from the multi-view or monocular videos~\cite{wang2022arah,peng2021animatable,guo2023vid2avatar}. These methods typically utilize a neural signed distance function (SDF) to implicitly represent the geometry, and a rigid bone transformation of SMPL~\cite{loper2015smpl} to capture the dynamics of the human body. Our method directly uses the implementation from \cite{lin2024relightable} to obtain the explicit canonical body mesh. We apply isotropic remeshing to the extracted mesh using MeshLab~\cite{cignoni2008meshlab}, ensuring that the body mesh contains about 40K vertices. The vertex positions of the processed canonical mesh are defined as $\{ \mathbf{x}^i\}_{  i \in [1,N_v] }$.

After obtaining the canonical mesh, we assign a series of attributes on each mesh vertex. The basic Gaussian attributes include quaternion rotation $\mathbf{r} \in \mathbb{R}^4$ and scale $\mathbf{s} \in \mathbb{R}^3$. The material attributes, comprising albedo $\mathbf{a} \in \mathbb{R}^3$, roughness $\gamma \in \mathbb{R}$ and specular tint $p \in \mathbb{R}$, are used for computing shading color later. Similar to \cite{peng2021animatable}, we obtain the linear blend shape (LBS) weight attribute $\mathbf{w} \in \mathbb{R}^{24}$ through the barycentric interpolation of the weights of corresponding triangle vertices on the SMPL body. The normal attribute $\mathbf{n} \in \mathbb{R}^3$ can be directly computed based on the mesh. We note that only $\{\mathbf{r},\mathbf{s},\mathbf{a},\gamma,p\}$ are the trainable parameters. Since rigid bone transformation is insufficient to model the dynamic body, we further define displacement attribute to model the pose-dependent non-rigid deformation, which can be calculated by $\Delta \mathbf{x}=d(\mathbf{x},\theta)$, where $\Delta \mathbf{x}$ is the displacement attribute, $\theta \in \mathbb{R}^{72}$ is the input pose, and $d$ is a multilayer perceptron network (MLP). To model the shadow produced by body occlusion, we also incorporate the visibility attribute $\mathbf{v} \in \mathbb{R}^{512}$, which is introduced in Section~\ref{sec:visibility}.

Next, a set of Gaussians is randomly initialized on the canonical body mesh. Their positions $\{\mathbf{x}_g^i | \mathbf{x}_g \in \mathbb{R}^3 \}_{ i \in [1,N_g]}$ are also trainable, allowing the Gaussians to move freely within the space. For a sample Gaussian $\mathbf{x}_g^i$, we adopt a similar idea from \cite{wu2023nerf} and assign attributes to the Gaussian. These attributes are calculated by the weighted average of $K_g$ nearest neighbors from the canonical mesh vertices
\begin{equation}
*_g^i = \frac{ \sum_{j \in \mathcal{S}^i_g} u_j(\mathbf{x}^i) \cdot *^j }{ \sum_{j \in \mathcal{S}^i_g} u_j(\mathbf{x}^i) }. 
\label{eqn:interpolate}
\end{equation}
$*$ represents one of the attributes $\{\mathbf{r},\mathbf{s},\mathbf{n},\mathbf{a},\gamma,p,\mathbf{w},\Delta \mathbf{x},\mathbf{v} \}$. $*^j$ and $*_g^i$ denote the attributes of the $j$th vertex and the $i$th Gaussian, respectively. $\mathcal{S}^i_g$ contains the indices of the $K_g$ nearest vertices to the Gaussian $\mathbf{x}_g^i$. $u_j(\mathbf{x}^i) = 1/\| \mathbf{x}_g^i - \mathbf{x}^j \|_2$ is the interpolation weight. We find it's sufficient to model the spatially varying high-frequency attributes by directly interpolating vertices if the vertex density is high. We always set the opacity of the Gaussians to $1$, so there is no need to model the opacity attribute. 

To animate the body, given a pose $\theta$, we can compute all the Gaussian attributes by Eq~\ref{eqn:interpolate}. The Gaussian positions with displacements are computed by $\mathbf{\bar{x}}_g=\mathbf{x}_g+\Delta \mathbf{x}_g$. Then Gaussian position can be deformed to the posed space $\mathbf{\bar{x}}_{gd}$ using linear blend shape, based on Gaussian's LBS weight attribute $\mathbf{w}_g$. The normal and rotation attributes can be rotated accordingly. We define the deformed normal and rotation of the Gaussian as $\mathbf{n}_{gd}$, $\mathbf{r}_{gd}$.

For the environment map, we use the same design as \cite{wu2023nerf,chen2022relighting4d,xu2023relightable}, which define the map as light probes $L \in \mathbb{R}^{16\times32\times3}$ with 512 discrete area lights. Figure~\ref{fig:pipeline} displays the spherical format of the environment map. In the posed space, to calculate the shading of a sample Gaussian at a given position $\mathbf{\bar{x}}_{gd}$ viewed from a certain direction $w_{o}$, we integrate the rendering equation by explicitly summing up the discrete light probes, 
\begin{equation}
\label{eqn:render_equation}
\begin{aligned}
&L_{o}(  \mathbf{\bar{x}}_{gd} , w_{o}) = \\
&\sum_{k=1}^{512} L^k \cdot A^k \cdot R(w_{i}^k, w_{o}, \mathbf{n}_{gd}) \cdot \mathbf{v}_g[k] \cdot \max(0,(w_{i}^k \cdot \mathbf{n}_{gd})).
\end{aligned}
\end{equation}
$L_{o}(\mathbf{\bar{x}}_{gd}, w_{o})$ is the output radiance. $L^k$, $A^k$ and $w_{i}^k$ are the radiance strength, area, and incident direction of the $k$th light, respectively. $\mathbf{v}_g[k]$ is the $k$th element of Gaussian's visibility attribute, showing whether the Gaussian could be observed by the $k$th light. $R$ is the Bidirectional Reflectance Distribution Function (BRDF). We use a simplified version of Disney BRDF~\cite{burley2012physically} to represent our material. If the normal of the Gaussian is facing away from the camera direction, we set Gaussian's opacity to zero. We also apply gamma correction to the output radiance to get the shading color. 

Finally, we are able to render the posed human with the posed position $\mathbf{\bar{x}}_{gd}$, scale $\mathbf{s}_g$, rotation $\mathbf{r}_{gd}$ and shading color of the Gaussians. The Gaussian rasterizer of 3DGS takes the above attributes as input and outputs the image, denoted as $I_{render}$.

\subsection{Visibility Computation}
\label{sec:visibility}

The shadows produced by body self-occlusion can enhance the realism of relighting results. However, it's not trivial to model such shadows quickly. Previous methods use ray tracing~\cite{wang2023intrinsicavatar,li2024animatable}, Distance Field Soft Shadow (DFSS)~\cite{xu2023relightable} or pretrained model~\cite{chen2022relighting4d,lin2024relightable} to calculate the visibility from the body surface to the light to model shadows. But they all encounter efficiency issues because these methods require inferring networks or conducting spatial intersections many times. Instead, we propose to use mesh rasterization to compute the visibility.

We first define the visibility on a mesh vertex as a vector $\mathbf{v} \in \{0,1\}^{512}$, which indicates whether the vertex could be observed from 512 discrete incoming light directions $\{w_i^k\}_{k \in [1,512]}$. To calculate the visibility, for a posed body mesh, we conduct orthographic projection towards the 512 directions and rasterize the projected triangles with Nvdiffrast~\cite{laine2020modular}. Nvdiffrast produces 2D images, each pixel of which indicates the rasterized triangle index. If a triangle appears in the 2D image, the visibility of its three vertices is set to $1$. Therefore, given a light direction, we can obtain the visibility map of each vertex towards the light direction, denoted as $V \in \{0,1\}^{N_v}$. 

Directly applying the above method to compute the visibility creates noisy results, as illustrated in Figure~\ref{fig:ablation_visib_postprocess}, because Nvdiffrast cannot render all the triangles on the image when casting rays from the camera to triangles at a grazing angle. Therefore, we need to post-process the visibility map. Specifically, for a given sample light direction $w_i^k$, the visibility map at this direction $V^k \in \{0,1\}^{N_v}$ is processed by, 
\begin{equation}
\bar{V}^k = \mathsf{mean}(\mathsf{median}(\mathsf{median}(V^k))),
\end{equation}
where \textsf{mean} and \textsf{median} are filters defined on mesh vertices. For each vertex on the mesh, \textsf{mean} calculates the average visibility value of its surrounding vertices, and \textsf{median} calculates the median visibility value. We illustrate the filters as follows,
\begin{equation}
\begin{aligned}
\mathsf{mean}(V) &= \{ \nu^i= \frac{1}{ |\mathcal{S}^i_v| } \sum_{k\in \mathcal{S}^i_v} V[k]  \}_{i\in [1,N_v]} \\
\mathsf{median}(V) &= \{ \nu^i=\mathsf{median}(\{ V[k] \}_{k\in \mathcal{S}^i_v}) \}_{i \in [1,N_v]} ,\\
\end{aligned}
\end{equation}
where $\mathcal{S}^i_v$ is the indices of $K_v$ nearest vertices to vertex $\mathbf{x}^i$, $V[k]$ is the $k$th element of visibility map. The median filter can help remove noise and smooth the visibility boundaries. The mean filter converts the value of visibility into a floating point number between 0 and 1, which can create soft shadows produced by area light sources, as shown in Figure~\ref{fig:ablation_visib_postprocess}. 

Finally, we obtain the $i$th vertex visibility attribute as $\mathbf{v}^i = \{ \bar{V}^k [i] \}_{k \in [1,512]}$ , which will be converted to Gaussian visibility attribute. The Gaussian visibility attribute is used in Eq~\ref{eqn:render_equation}, as described in Section~\ref{sec:gaussianavatar}.

\emph{Discussion with shadow mapping.} While our method shares a similar idea with shadow mapping~\cite{williams1978casting} in rasterizing meshes from the light source for fast shadow calculation, there is still a difference. Shadow mapping rasterizes a depth map from the light source and conducts depth testing from viewpoints to check the visibility of a point on the mesh, while our method rasterizes from the light source to obtain a triangle's visibility and directly uses triangle's visibility to approximate a point's visibility.

\begin{table*}[t]
\caption{Quantitative comparison. We compare with baselines on SyntheticDataset. As we use the entire image to calculate metrics, the results will be higher than those reported in past works~\cite{xu2023relightable}. All methods are trained and rendered at a resolution of 500$\times$500.}
\vspace{-3mm}
\begin{tabularx}{\textwidth}{l|XXXXXXXXXX}
\toprule
\multirow{2}{*}{Method} & \multicolumn{1}{c}{Normal} & \multicolumn{3}{c}{Albedo} & \multicolumn{3}{c}{Relighting (Training poses)} & \multicolumn{3}{c}{Relighting (Novel poses)} \\
       & Error $\downarrow$ & PSNR $\uparrow$    & SSIM $\uparrow$   & LPIPS $\downarrow$  & PSNR $\uparrow$    & SSIM $\uparrow$   & LPIPS $\downarrow$  & PSNR $\uparrow$    & SSIM $\uparrow$   & LPIPS $\downarrow$  \\ \hline
Ours   & \cello{9.322} & \cello{34.3489} & \cello{0.9428} & \cello{0.1695} & \cello{36.1635} & \cello{0.8640} & \cello{0.0555} & \cello{28.8396} & \cello{0.8308} & \cello{0.0737} \\
RA-Lin~\cite{lin2024relightable}   & 10.155 &30.6854 & 0.9342 & 0.1766 & 33.1736 & 0.8548 & 0.0634 & 28.4573 & 0.8281 & 0.0774 \\
MeshAvatar~\cite{chen2024meshavatar}   & 11.685 &30.5407 & 0.9283 & 0.1859 & 33.4034 & 0.8522 & 0.0724 & 27.9275 & 0.8217 & 0.0878 \\
RA-Xu~\cite{xu2023relightable}  & 10.132 & 32.0395 & 0.9365 & 0.1790 & 33.5669 & 0.8544 & 0.0675 & 28.3756 & 0.8258 &  0.0817 \\
IA~\cite{wang2023intrinsicavatar}  & 11.774 &27.9200 & 0.9252 & 0.1839 & 31.5716 & 0.8463 & 0.0693 & 28.1717 & 0.8259 & 0.0807 \\
R4D~\cite{chen2022relighting4d} & 24.502 &24.9638 & 0.9026 & 0.2050 & 28.9139 & 0.8306 & 0.0879 & 25.4990 & 0.8081 & 0.1024 \\
\bottomrule
\end{tabularx}
\label{table:compare}
\end{table*}

\subsection{Training}
\label{sec:training}

Based on the above Gaussian representation, we can combine light probes to train the model under a given pose and viewpoint. We first apply the same image loss as 3DGS:
\begin{equation}
\mathcal{L}_{img} = (1-\lambda_{img}) \mathcal{L}_1 + \lambda_{img} \mathcal{L}_{D-SSIM},
\label{eqn:image_loss}
\end{equation}
where $\lambda_{img}=0.2$. There may be ambiguities in solving materials and lighting~\cite{chen2022relighting4d,lin2024relightable,xu2023relightable}, so we also apply some regularization. We add a smooth loss on the vertex material attributes:
\begin{equation}
\mathcal{L}_{smooth} = \sum_{i=1}^{N_v} \sum_{k \in \mathcal{S}^i_v} \| *^i - *^k \|_1,
\end{equation}
where $*$ represents one of the material attributes $\{ \mathbf{a},\gamma, p \}$, $\mathcal{S}^i_v$ is the indices of $K_v$ nearest vertices to vertex $\mathbf{x}^i$. To prevent the Gaussians from straying too far from the mesh surface, we add a mesh distance loss:
\begin{equation}
\mathcal{L}_{mdist} = \sum_{i=1}^{N_g} \sum_{k \in \mathcal{S}^i_g} \| \mathbf{x}_g^i - \mathbf{x}^k \|_2, 
\label{eqn:mesh_dist_loss}
\end{equation}
where $\mathcal{S}^i_g$ contains the indices of $K_g$ nearest vertices to the Gaussian $\mathbf{x}_g^i$. We also add the displacement regularization on the vertex attribute:
\begin{equation}
\mathcal{L}_{disp} = \sum_{i=1}^{N_v} \| \Delta \mathbf{x}^i \|_2.
\end{equation}

With the above training process, we can recover the materials and the environment map. To relight the body under novel lighting, we simply replace the trained light probes $L$ with a new environment map. However, under new lighting conditions, the human body produces undesirable results, as shown in Figure~\ref{fig:ablation_scale}. This is because some Gaussians are too long or too large, so that the normal is not accurate compared to the range covered by the Gaussians. Therefore, we apply Gaussian scale loss to prevent Gaussian's scale from growing too large:
\begin{equation}
\mathcal{L}_{scale} = \sum_{i=1}^{N_g} \max(0, \mathbf{s}_g^i - s_0),
\label{eqn:scale_loss}
\end{equation}
where $s_0=0.005$ is the scale threshold. 

Together we have all the training loss as
\begin{equation}
\begin{aligned}
\mathcal{L} =& \mathcal{L}_{img} + \lambda_{smooth} \mathcal{L}_{smooth} \\
&+ \lambda_{mdist} \mathcal{L}_{mdist} + \lambda_{disp} \mathcal{L}_{disp} + \lambda_{scale} \mathcal{L}_{scale}.
\end{aligned}
\end{equation}
We set $\lambda_{smooth}=0.002$, $\lambda_{mdist}=0.1$, $\lambda_{disp}=0.02$, $\lambda_{scale}=10$ for all the experiments. Since the detected poses may be noisy, we also optimize the input poses during training. The poses are only optimized during the stage where Gaussians are deformed to the posed space. In other stages, we stop the gradient of the poses. Figure~\ref{fig:pipeline} shows when the poses serve as learnable parameters.

During training, we conduct 3DGS's densification method to increase the number of Gaussians. We find that if some parts of the body are seldom seen during training, there will be too few Gaussians in these areas, resulting in holes under new poses (See Figure~\ref{fig:ablation_compare}). To solve this problem, in addition to the original densification method, we propose a new method to increase the number of Gaussians based on the density of Gaussians on the mesh surface. Specifically, we count the number of Gaussians around each vertex based on the $K_g$ nearest neighbors $\mathcal{S}^i_g$ and further estimate the Gaussian density of each triangle on the mesh. If the Gaussian density of a triangle is below a certain density threshold, we randomly add Gaussians to that triangle with a certain probability. The lower the Gaussian density of the triangle, the greater this probability will be. Please see the supplementary materials for more details.

\subsection{Appearance Editing}
\label{sec:appedit}

Given a reconstructed relightable avatar, we can edit the albedo to change the appearance. For the editing paradigm, we render an albedo map from a given viewpoint. The albedo map can be edited, and the appearance of the body changes accordingly. We achieve this goal through single-view attributes  optimization. Specifically, we first determine which vertices and Gaussians are within the mask range of the edited area. We only optimize the Gaussian and vertex attributes within the mask range. Then we apply image loss as Eq~\ref{eqn:image_loss} on the edited albedo and rendered albedo. We also impose the mesh distance loss (Eq~\ref{eqn:mesh_dist_loss}) and scale loss (Eq~\ref{eqn:scale_loss}). Note that in the above process, only the Gaussian attribute $\mathbf{x}_g$ and the vertex attributes $\mathbf{r},\mathbf{s},\mathbf{a}$ are optimizable. $\gamma, p$ and pose-dependent MLP are not optimized.

\begin{wrapfigure}{r}{0.26\textwidth}
\centering\includegraphics[width=0.26\textwidth]{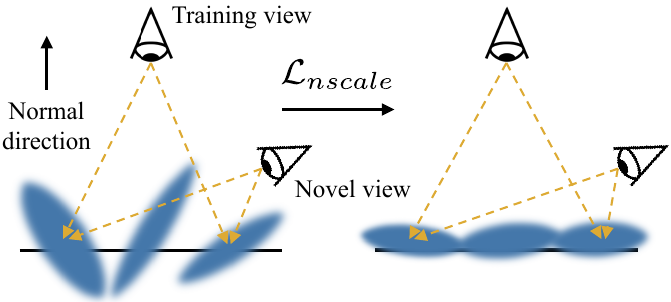}
\end{wrapfigure}
Since we only optimize from single viewpoint, the Gaussians may grow towards normal direction during training, producing blending artifacts under novel viewpoints (See inset figure and Figure~\ref{fig:ablation_nscale}). So we add a normal scale loss to make the Gaussian's scale towards the normal direction as small as possible to avoid the blending artifacts,
\begin{equation}
\mathcal{L}_{nscale} = \sum_{i \in \mathcal{E}_g} \max(0, \| \mathsf{Rot}(\mathsf{Inv}(\mathbf{r}^i_g), \mathbf{n}^i_g) \odot \mathbf{s}^i_g \|_2 - s_n),
\end{equation}
where $\mathcal{E}_g$ is the indices of the Gaussians within the mask, $\mathsf{Inv}$ returns the inverse rotation, $\mathsf{Rot}({\mathbf{r}, \mathbf{n})}$ applies rotation $\mathbf{r}$ on vector $\mathbf{n}$, $\odot$ is element-wise multiplication of vectors, and $s_n=0.001$ is the scale threshold.

\begin{figure*}[tbp]
  \begin{center}
    \includegraphics[width=0.99\textwidth]{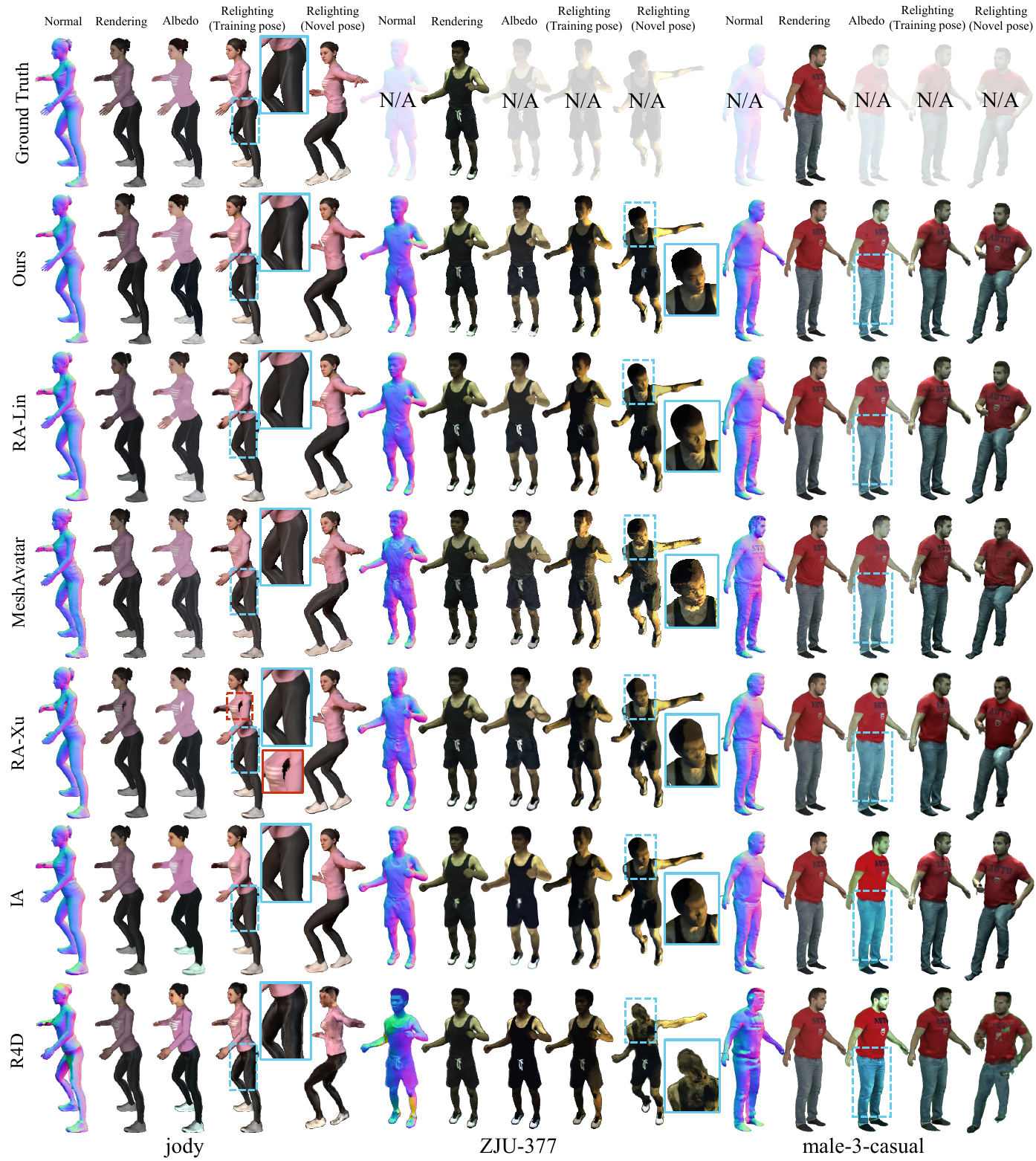}
  \end{center}
  \vspace{-5mm}
  \caption{Qualitative comparison with RA-Lin~\cite{lin2024relightable}, MeshAvatar~\cite{chen2024meshavatar}, RA-Xu~\cite{xu2023relightable}, IA~\cite{wang2023intrinsicavatar} and R4D~\cite{chen2022relighting4d}. We show the albedo and the relighting results under training and novel poses on both synthetic data (jody, rendered at test viewpoints) and real data (ZJU-377 and male-3-casual, rendered at training viewpoints). Compared to the baselines, our method can achieve finer body details (jody's leggings, ZJU-377's face and male-3-casual's jeans) and the specular effects that are closest to the ground truth (jody's leggings).
  }
  \label{fig:compare}
\end{figure*}

\subsection{Implementation Details}
\label{sec:impldetails}

For the detailed design, the displacement MLP has four layers, each with a width of 256 and activated by ReLU. We adopt Random Fourier Features~\cite{tancik2020fourier} to let the displacement network learn high frequency coordinate signals. For the calculation of K nearest neighbors, we set $K_g=3$ and $K_v=19$ in our experiments. We use a fast KNN implementation\footnote{https://github.com/lxxue/FRNN}, which can calculate KNN 500 times per second. Before training, we pre-calculate the visibility of all the training poses to avoid spending too much time on computing the visibility on the fly. 

At the begining of the training, we randomly initialize 7K Gaussians on the mesh surface. The increase in the number of Gaussians relies on both 3DGS's densification and our densification methods, as illustrated in Section~\ref{sec:training}. As for our densification method, it is performed every 100 iterations, starts at 10K iterations and ends at 25K iterations. It takes 30K iterations to train a relightable avatar, which contains about 100K Gaussians.

After training, the trained avatar can be relit by replacing the trained environment map $L$ with a novel environment map $L'$ and animated with novel poses. The visibility is computed on the fly during testing. In order to make the model generalize better to novel poses, for a vertex $\mathbf{x}^i$, we calculate the average displacement $\Delta \mathbf{\tilde{x}}^i = \frac{1}{T}\sum_{t=1}^{T} d(\mathbf{x}^i, \theta^t)$ across all the training poses. During novel pose testing, we use the average displacement $\Delta \mathbf{\tilde{x}}^i$ to replace the vertex's original displacement attribute $\Delta \mathbf{x}^i$. 

For appearance editing, we set the learning rate of Gaussian's position attribute to 1.6e-6. We only use 3DGS's densification method during optimization and set the densification gradient threshold to 1e-4. We train 3K iterations for each editing.



\section{Evaluation}

In this section, we conduct experiments on multiple datasets (Section~\ref{sec:datasets}) and introduce the metrics to validate the results (Section~\ref{sec:metrics}). Our approach outperforms baselines in achieving human body relighting (Section~\ref{sec:comparison}). We conduct ablation studies that validate the effectiveness of several designs (Section~\ref{sec:ablation}). We also show the appearance editing ability of our method (Section~\ref{sec:appeditexp}). In terms of rendering efficiency, our method surpasses previous works, achieving rendering at interactive frame rates (Section~\ref{sec:speed}). Additionally, we show the interactive relighting results in supplementary video.

\begin{figure*}[t]
  \begin{center}
    \includegraphics[width=0.75\textwidth]{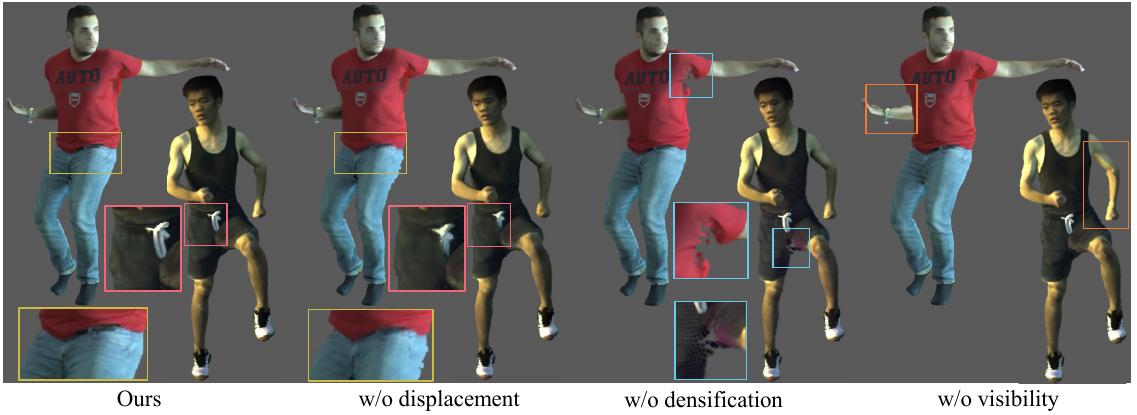}
  \end{center}
  \vspace{-5mm}
  \caption{Ablation study on non-rigid displacement, our densification and visibility. All results are rendered under novel poses and new environment light.
  }
  \label{fig:ablation_compare}
\end{figure*}

\begin{table*}[]
\caption{Ablation study on several designs. All methods are trained and rendered at a resolution of 1K$\times$1K.}
\vspace{-3mm}
\begin{tabularx}{\textwidth}{l|XXXXXXXXX}
\toprule
\multirow{2}{*}{Method} & \multicolumn{3}{c}{Albedo} & \multicolumn{3}{c}{Relighting (Training poses)} & \multicolumn{3}{c}{Relighting (Novel poses)} \\
                          & PSNR $\uparrow$    & SSIM $\uparrow$   & LPIPS $\downarrow$  & PSNR $\uparrow$    & SSIM $\uparrow$   & LPIPS $\downarrow$  & PSNR $\uparrow$    & SSIM $\uparrow$   & LPIPS $\downarrow$  \\ \hline
Ours                      & \celly{34.3398} & \celly{0.9580} & \celly{0.1985} & \cello{36.1024} & \cello{0.8826} & \cello{0.1647} & 28.3886 & \celly{0.8546} & \cello{0.1815} \\
w/o visibility            & \cello{34.3765} & \cello{0.9584} & \cello{0.1981} & 35.3971 & 0.8820 & 0.1652 & 28.2869 & 0.8541 & 0.1819 \\
w/o our densification         & 34.2508 & 0.9573 & 0.2020 & 35.8359 & 0.8807 & 0.1717 & \celly{28.4261} & 0.8536 & 0.1866 \\
w/o displacement & 30.1185 & 0.9501 & 0.2059 & 32.7894 & 0.8743 & 0.1735 & \cello{28.4539} & \cello{0.8552} & 0.1854 \\
w/o $\mathcal{L}_{scale}$  & 34.0571 & 0.9575 & 0.1991 & \celly{35.9361} & \celly{0.8822} & \celly{0.1650} & 28.3980 & \celly{0.8546} & \celly{0.1816} \\
w/ SMPL mesh              & 32.4590 & 0.9539 & 0.2031 & 34.0575 & 0.8777 & 0.1716 & 27.8660 & 0.8521 & 0.1858 \\ \bottomrule
\end{tabularx}
\label{table:ablation}
\end{table*}

\subsection{Datasets}
\label{sec:datasets}

We use both synthetic data and real data for validation.

\textbf{SyntheticDataset} To quantitatively validate the methods, we follow~\cite{lin2024relightable} to create a synthetic dataset. We use two models from Mixamo\footnote{https://www.mixamo.com}, transfer the motions from ZJUMoCap~\cite{peng2021neural} onto the models, and render images from multiple views. We render 4 viewpoints for training, and another 4 viewpoints for novel view evaluation. Each sequence contains 100 frames. We use an even distribution of 10 frames from each sequence for training pose evaluation. We also render images with new lighting and novel poses for testing under new lighting and poses.

\textbf{ZJUMoCap}~\cite{peng2021neural} Each sequence of the dataset contains a person captured from 23 different viewpoints in a light stage with unknown illumination. For each sequence, we uniformly select 4 viewpoints and 300 frames for training.

\textbf{PeopleSnapshot}~\cite{alldieck2018video} The dataset includes a person turning around in front of a single camera in an A-pose. We use 300 frames from each sequence for training.

\subsection{Metrics}
\label{sec:metrics}

For quantitative evaluations, we use Peak Signal-to-Noise Ratio (PSNR), Structural Similarity Index Measure (SSIM), and Learned Perceptual Image Patch Similarity (LPIPS)\cite{zhang2018unreasonable} as metrics. We also follow IA~\cite{wang2023intrinsicavatar} and RA-Xu~\cite{xu2023relightable} to compute the normal difference (in degrees) between the results and the ground truth. On synthetic data, we render the albedo and the relighting results from both training poses and novel poses for metric computation. Our method can render at high resolution (1K $\times$ 1K). However, for fair comparison, in Section~\ref{sec:comparison} and Section~\ref{sec:speed}, all the comparison experiments are trained and rendered at a resolution of 500 $\times$ 500. In other experiments, we train and test our method at a resolution of 1K. Our metrics are calculated on the entire image, including the black background. Similar to \cite{wang2023intrinsicavatar}, we compute per-channel scaling factor to align the albedo and rendered images with the ground truth, addressing the ambiguity issue of inverse rendering between different methods.

For the real-world dataset, we primarily present the qualitative results under novel poses and illuminations for evaluations.

\subsection{Comparison}
\label{sec:comparison}

\textbf{Baselines} We compare our work with RelightableAvatar-Lin (RA-lin~\cite{lin2024relightable}), RelightableAvatar-Xu (RA-Xu~\cite{xu2023relightable}), IntrinsicAvatar (IA~\cite{wang2023intrinsicavatar}), Relighting4D (R4D~\cite{chen2022relighting4d}) and MeshAvatar~\cite{chen2024meshavatar}. To ensure fair comparison, we compare our method where the data and source code are available. Some baselines~\cite{chen2022relighting4d,wang2023intrinsicavatar} are originally designed for monocular videos. We adapt them for a multi-view setting. Other baselines such as \cite{li2024animatable,bolanos2024gaussian,sun2023neural,iqbal2023rana, chen2024meshavatar} are not employed due to the lack of data or code. To exclude the effect of pose correction for a fair comparison, we use the optimized poses from our method, and disable the pose correction stage for all methods during training.

\textbf{Results} Figure~\ref{fig:compare} shows the comparison. We present the results under both synthetic and real datasets. The images are all adjusted by per-channel scaling to make the brightness similar for different methods. R4D~\cite{chen2022relighting4d} assigns latent codes to each frame to model the appearance, therefore the reconstructed human body naturally fails to generalize to novel poses. For IA~\cite{wang2023intrinsicavatar}, the results appear relatively blurry, because IA doesn't model pose-dependent non-rigid deformation for the animated human. MeshAvatar~\cite{chen2024meshavatar} reconstructs a non-smooth body surface under the sparse-view setting, resulting in noisy relighting outcomes. For RA-Xu~\cite{xu2023relightable}, its design of the signed distance function tends to create hollow artifacts under the armpits (Figure~\ref{fig:compare}, red box). Overall, although our results seem roughly similar to those of RA-Lin~\cite{lin2024relightable} and RA-Xu~\cite{xu2023relightable}, our results perform better in preserving the details. Compared with RA-Lin and RA-Xu, our method achieves clearer details on the stripes of jody's leggings, the face of ZJU-377 and the jeans of male-3-casual (Figure~\ref{fig:compare}, blue box). Our method also produces specular results closer to ground truth on jody's leggings. Table~\ref{table:compare} shows the quantitative results between our method and the baselines. Our method outperforms other methods under various metrics. We note that in the novel pose test, due to some misalignments between the posed human body and the ground truth, there will be a noticeable drop in metrics, as illustrated in \cite{lin2024relightable} as well. The comparison results show that our method can better recover materials and render superior relighting results.

\subsection{Ablation Study}
\label{sec:ablation}

We conduct ablation studies on several components of our method, including without incorporating visibility, without applying our densification method, without displacement, without adding Gaussian scale loss, and using the SMPL mesh instead of the SDF reconstructed mesh. We train and render the images at a resolution of 1K in this section. We conduct the quantitative experiments on synthetic dataset, which is presented in Table~\ref{table:ablation}. We also show qualitative results of real data under novel poses and lighting in Figure~\ref{fig:ablation_compare}.

\textbf{Visibility} In our results, there are shadows on the arm occluded by the body. In Figure~\ref{fig:ablation_compare}, without visibility, the arms still reflect the light even when occluded by the body, resulting in less realistic rendering results.

\textbf{Our densification} For the case of male-3-casual, the character maintains an A-pose in the training data and the underarm area is not visible to the camera for all the frames. As a result, Gaussians barely move and the view-space positional gradients are small, thus the original 3DGS's densification doesn't increase the number of Gaussians under the arm. We add Gaussians to areas where the Gaussian density is low, such as the armpits, to prevent holes when the avatar is animated under a novel pose (See Figure~\ref{fig:ablation_compare}).

\textbf{Displacement} In Figure~\ref{fig:ablation_compare}, without displacement, it is hard for the model to capture the appearance differences across different poses, resulting in blurry results. Some details, like the strings on the pants, are also difficult to be accurately reconstructed.

\begin{figure}[tbp]
  \begin{center}
    \includegraphics[width=0.490\textwidth]{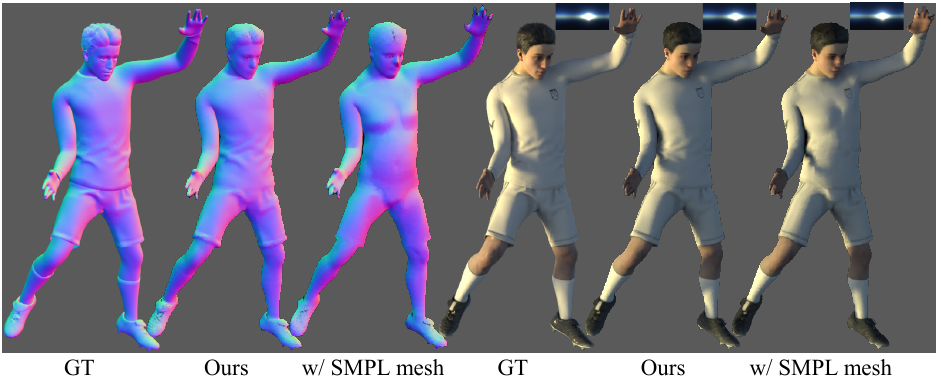}
  \end{center}
  \vspace{-5mm}
  \caption{Ablation study on using SMPL mesh. We present the normal and relighting results using the SMPL mesh and SDF mesh. With SMPL mesh, Gaussians may access wrong normal attributes, resulting in inaccurate relighting results.}
  \label{fig:ablation_sdfmesh}
\end{figure}

\begin{figure}[tbp]
  \begin{center}
    \includegraphics[width=0.420\textwidth]{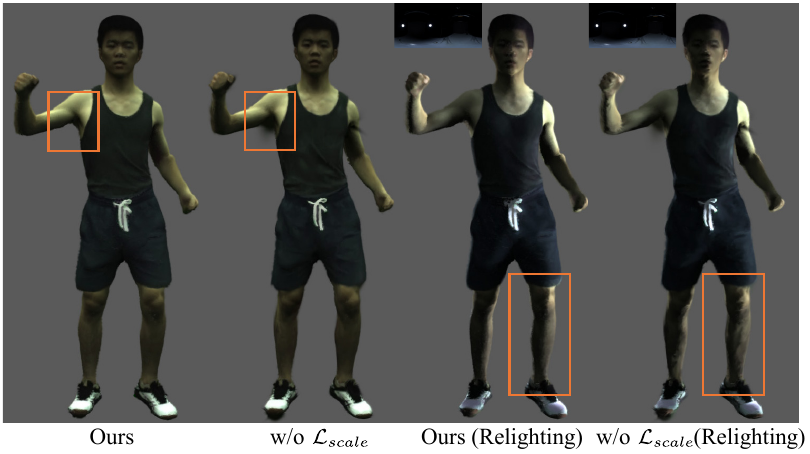}
  \end{center}
  \vspace{-5mm}
  \caption{Ablation study on scale loss. We present the rendering and relighting results from a novel viewpoint. Without scale loss, The large Gaussians create artifacts under the arms and produce a scaly appearance on the leg under novel lighting.}
  \label{fig:ablation_scale}
\end{figure}

\begin{figure}[tbp]
  \begin{center}
    \includegraphics[width=0.490\textwidth]{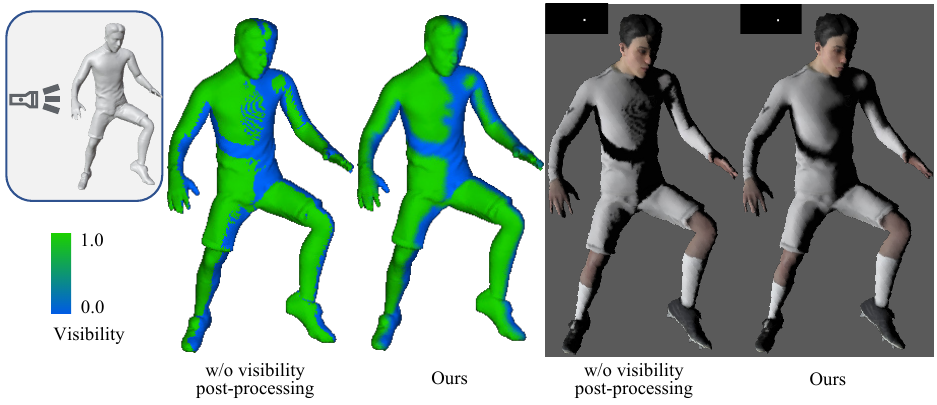}
  \end{center}
  \vspace{-5mm}
  \caption{Ablation study on visibility post-processing. The post-processed visibility map creates pleasant and soft shadow effects.}
  \label{fig:ablation_visib_postprocess}
\end{figure}

\textbf{Using SMPL mesh} For fairness, we also perform remeshing on the SMPL mesh to obtain a mesh with approximately 40K vertices, enabling the vertex attributes to be high-frequency. The SMPL mesh may differ significantly from the actual body geometry, causing the normal attributes obtained by the Gaussians to be inaccurate. Figure~\ref{fig:ablation_sdfmesh} shows the relighting results with the normal from the SMPL mesh. Compared to the ground truth, it fails to present the shadows caused by clothing's wrinkles and casts incorrect shadows on the chest.

\textbf{No Gaussian scale loss} Figure~\ref{fig:ablation_scale} shows the results without the constraint on scale attribute. Since it is difficult to see the underarm area in the training viewpoints, the Gaussians in that part can grow relatively large without any constraint. In novel viewpoints, these Gaussians create artifacts. Furthermore, the relighting results may produce a scaly appearance. This is because the normal of a Gaussian is interpolated only from the Gaussian's center. However, when the Gaussian is large, this normal is not accurate enough to cover the entire range of the Gaussian.

We also demonstrate the importance of visibility post-processing. In Figure~\ref{fig:ablation_visib_postprocess}, we cast the light source from the left side to light the mesh and calculate the visibility map. Without post-processing, the visibility map calculated by the mesh rasterization process produces noise, such as the chest, leading to poor relighting results. Our post-processing not only smooths the boundary of the visibility map, but also produces the soft shadow effect that can be generated by an area light.

\begin{figure}[!tbp]
  \begin{center}
    \includegraphics[width=0.490\textwidth]{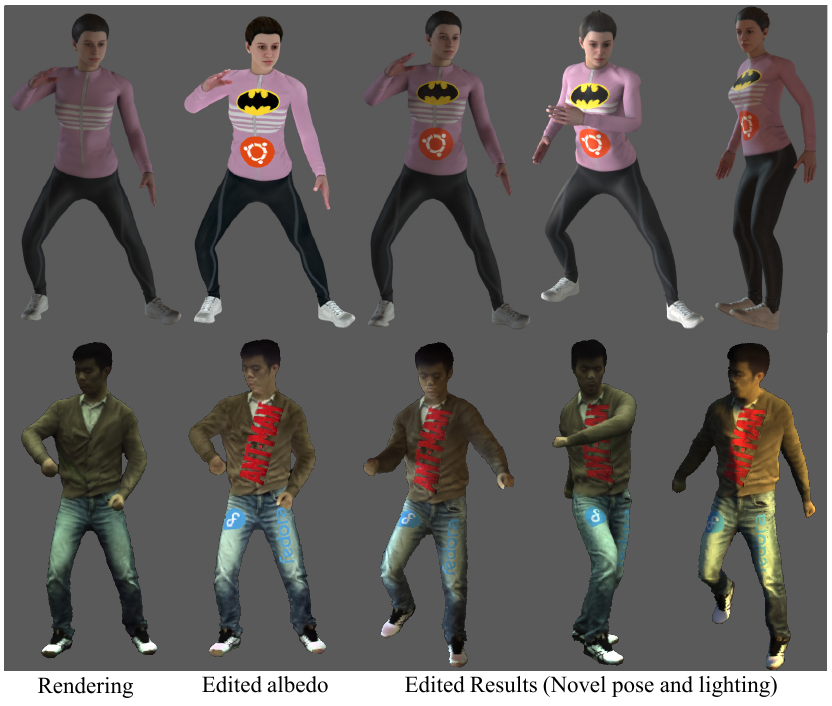}
  \end{center}
  \vspace{-5mm}
  \caption{Appearance editing results.}
  \label{fig:edit_texture}
\end{figure}

\begin{figure}[!tbp]
  \begin{center}
    \includegraphics[width=0.490\textwidth]{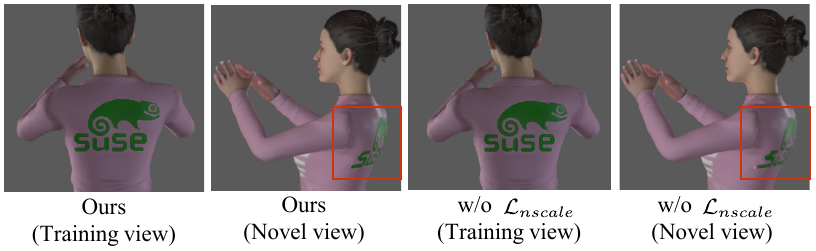}
  \end{center}
  \vspace{-5mm}
  \caption{Ablation study on $\mathcal{L}_{nscale}$.}
  \label{fig:ablation_nscale}
\end{figure}

\subsection{Appearance Editing}
\label{sec:appeditexp}

Figure~\ref{fig:edit_texture} shows the rendering images, edited albedo and the edited relighting results under novel poses. Even if editing and optimizing from a single viewpoint, our editing method can generalize to novel viewpoints and poses. Figure~\ref{fig:ablation_nscale} also shows the effectiveness of normal scale loss. When observing the edited albedo from a grazing view, the results may be bad due to the growth of Gaussian towards the normal direction. Our normal scale loss makes the Gaussians flat, making the albedo still complete under novel viewpoints.

\subsection{Speed}
\label{sec:speed}

\begin{table}[!tbp]
\caption{Time comparison with baselines.}
\vspace{-3mm}
\resizebox{\columnwidth}{!}{
\begin{tabular}{c|cccccc}
\toprule
         & RA-Lin~\cite{lin2024relightable} & MeshAvatar~\cite{chen2024meshavatar}
         & RA-Xu~\cite{xu2023relightable} & IA~\cite{wang2023intrinsicavatar}   & R4D~\cite{chen2022relighting4d}  & Ours                            \\ \hline
Training & 44h    &  8.6h & 33h   & 3.5h & 49h  & 55min (+6h, SDF mesh)           \\
Render   & 15.0s  &  0.180s & 4.9s  & 5.5s & 4.0s & 0.145s (0.021s, no visibility) \\ \bottomrule
\end{tabular}}
\label{table:time_compare}
\end{table}

\begin{table}[!tbp]
\caption{Time for each part to render 100 images.}
\vspace{-3mm}
\centering
\resizebox{0.9\columnwidth}{!}{
\begin{tabular}{cccc|c}
\toprule
KNN   & Visibility & Shading & GS renderer & Total  \\ \hline
0.093s & 12.107s     & 1.892s   & 0.180s     & 14.272s \\ \bottomrule
\end{tabular}}
\label{table:module_time}
\end{table} 

We also test the training and rendering time of our method. All methods are trained on a single NVIDIA RTX3090 GPU and rendered on a NVIDIA RTX4090 GPU. Table~\ref{table:time_compare} shows the time for different methods to train a model and render one image. Our method takes 6 hours to obtain the mesh, and about one hour to train. We use MLP-based SDF~\cite{lin2024relightable} to obtain the mesh, but it can be further accelerated by employing methods based on iNGP~\cite{muller2022instant}.

Table~\ref{table:module_time} also shows the time each part takes during the rendering process of our method. Even though the calculation of visibility takes a relatively large proportion of time, our method is still efficient enough for rendering at 6.9 fps, or 47.6 fps without visibility. We also emphasize that our method does not require any preprocessing or pretraining for novel environment map or poses to achieve such speed, allowing us to switch the lighting and animate the body arbitrarily. Please refer to the supplementary video for more interactive results.

\subsection{Limitation and Future Work}

Similar to previous methods~\cite{lin2024relightable,xu2023relightable,wang2023intrinsicavatar}, our method cannot model pose-dependent wrinkles because the material properties (albedo, roughness, specular tint) do not change with the poses, and we plan to model pose-dependent materials in the future. Our method also cannot reconstruct loose clothing, like dress. Using physics-based simulation to model the geometry of the clothing and optimizing the Gaussians based on the simulated geometry may resolve this issue, which we will leave for future work. Our method uses \cite{lin2024relightable} to obtain body mesh, which utilizes MLP to model the SDF and is slow to train. Some methods based on iNGP~\cite{muller2022instant} could acquire the mesh faster.


\section{Conclusion}

Given sparse-view or monocular videos of a person under unknown illumination, we can create a relightable and animatable human body. Thanks to the rendering framework of 3DGS, attribute acquisition via K nearest neighbors, and visibility calculation based on mesh rasterization, our method achieves high-quality relighting and interactive rendering speeds, enabling broader applications of digital humans and virtual reality. 

\section*{Acknowledgment}
The authors would like to thank reviewers for their insightful comments. This work is supported by the National Key Research and Development Program of China (No.2022YFF0902302), NSF China (No. 62322209 and No. 62421003), the gift from Adobe Research, the XPLORER PRIZE, and the 100 Talents Program of Zhejiang University.


\ifCLASSOPTIONcaptionsoff
  \newpage
\fi

\bibliographystyle{IEEEtran}
\bibliography{TVCG-2024-07-0555_Bib}

\begin{IEEEbiography}[{\includegraphics[width=1in,height=1.25in,clip,keepaspectratio]{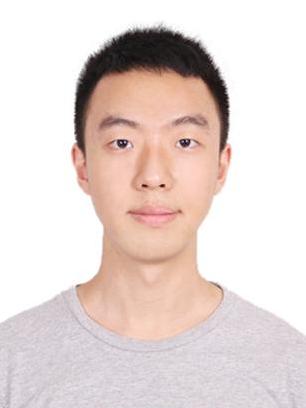}}]{Youyi Zhan} is working toward the Ph.D. degree at the State Key Lab of CAD\&CG, Zhejiang University. His research interests include deep learning, image processing and garment animation.
\end{IEEEbiography}

\begin{IEEEbiography}[{\includegraphics[width=1in,height=1.25in,clip,keepaspectratio]{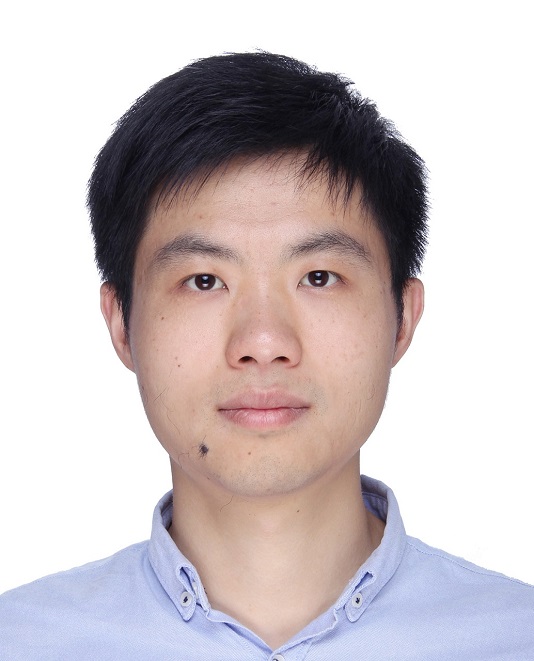}}]{Tianjia Shao} received his BS from the Department of Automation, and his PhD in computer science from Institute for Advanced Study, both in Tsinghua University. He is currently a ZJU100 Young Professor in the State Key Laboratory of CAD\&CG, Zhejiang University. Previously he was an Assistant Professor (Lecturer in UK) in the School of Computing, University of Leeds, UK. His current research focuses on 3D scene reconstruction, digital human creation, and 3D AIGC.
\end{IEEEbiography}

\begin{IEEEbiography}[{\includegraphics[width=1in,height=1.25in,clip,keepaspectratio]{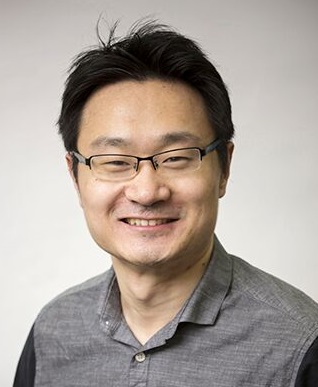}}]{He Wang}
received his PhD and did a post-doc in the School of Informatics, University of Edinburgh and his BS in Zhejiang University, China. He is an Associate Professor in the Virtual Environment and Computer Graphics (VECG) group, at the Department of Computer Science, University College London and a Visiting Professor at the University of Leeds. He is also a Turing Fellow and an Academic Advisor at the Commonwealth Scholarship Council. He serves as an Associate Editor of Computer Graphics Forum. His current research interest is mainly in computer graphics, vision and machine learning. Previously he was an Associate Professor and Lecturer at the University of Leeds UK and a Senior Research Associate at Disney Research Los Angeles.
\end{IEEEbiography}

\begin{IEEEbiography}[{\includegraphics[width=1in,height=1.25in,clip,keepaspectratio]{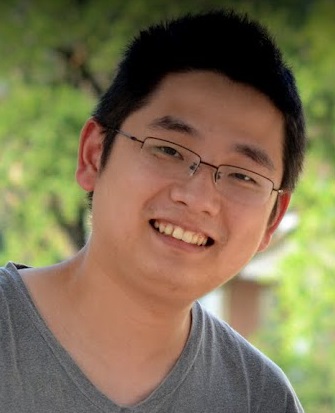}}]{Yin Yang}
received the PhD degree in computer science from the University of Texas at Dallas, in 2013. He is an associate professor with Kahlert School of Computing, University of Utah. He co-direct Utah Graphics Lab with his colleague Prof. Cem Yuksel. He is also affiliated with Utah Robotic Center. Before that, He was a faculty member at University of New Mexico and Clemson University. His research aims to develop efficient and customized computing methods for challenging problems in Graphics, Simulation, Deep Learning, Vision, Robotics, and many other applied areas.
\end{IEEEbiography}

\begin{IEEEbiography}[{\includegraphics[width=1in,height=1.25in,clip,keepaspectratio]{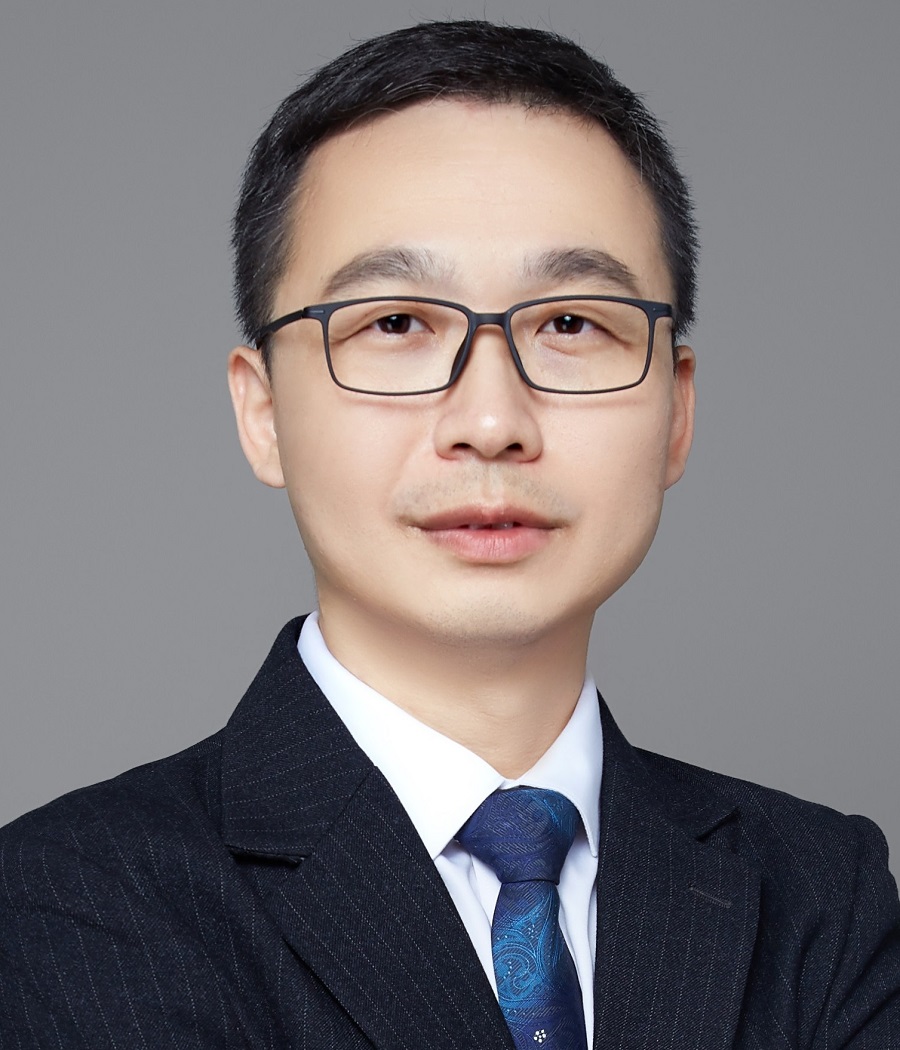}}]{Kun Zhou}
received the BS degree and PhD degree in computer science from Zhejiang University, in 1997 and 2002, respectively. He is a Cheung Kong professor with the Computer Science Department, Zhejiang University, and the director of the State Key Lab of CAD\&CG. Prior to joining Zhejiang University in 2008, he
was a leader researcher of the Internet Graphics Group, Microsoft Research Asia. He was named one of the world’s top 35 young innovators by MIT Technology Review in 2011, was elected an IEEE Fellow in 2015 and an ACM Fellow in 2020. His research interests are in visual computing, parallel computing, human computer interaction, and virtual reality.
\end{IEEEbiography}

\clearpage

\appendices
\section{BRDF Definition}

We use a simplified Disney BRDF~\cite{karis2013real} in our material model, and introduce specular tint to model situations where the surface has fewer specular components. The BRDF function $R(w_i,w_o, \mathbf{n})$ takes light incoming and outgoing direction $w_i, w_o$, normal $\mathbf{n}$, per-channel albedo $a$, roughness $\gamma$ and specular tint $p$ as input. We omit the material properties to simplify the expression. BRDF is defined as
\begin{equation}
R(w_i,w_o, \mathbf{n}) = 
\frac{a}{\pi} + p \cdot \frac{ D(w_h, \mathbf{n}) F(w_o, w_h) G(w_o, w_i, \mathbf{n}) }{4(\mathbf{n} \cdot w_i ) (\mathbf{n} \cdot w_o)},    
\end{equation}
where $w_h=\frac{w_o + w_i}{\| w_o + w_i \|}$ is the half vector between $w_i$ and $w_o$. $D$ is the normal distribution function, $F$ is the Fresnel reflection and $G$ is the geometric attenuation or shadowing factor. We use the same terms as \cite{karis2013real}, which are defined as
\begin{align}
&D(w_h,\mathbf{n}) = \frac{ \alpha^2 }{ \pi( (\mathbf{n} \cdot w_h)^2 ( \alpha^2 - 1 ) +1 )^2 } \quad \mathrm{s.t.} \ \alpha=\gamma^2 \\
&F(w_o,w_h)= \notag \\ 
&\qquad F_0 + (1-F_0)2^{(-5.55473(w_o \cdot w_h) - 6.98316)(w_o \cdot w_h)} \\
&G(w_o, w_i, \mathbf{n}) = G_1(w_i)G_1(w_o) \\
& \qquad \begin{aligned}
\mathrm{s.t.} \quad G_1(w) &= \frac{ \mathbf{n} \cdot w }{ (\mathbf{n} \cdot w)(1-k) + k } \\
k &= \frac{(\gamma+1)^2}{8}, 
\end{aligned} \notag
\end{align}
where $F_0=0.04$ is a fixed Fresnel value. 

\section{Gaussian Densification Algorithm}

The densification method of 3DGS determines whether to split or clone Gaussians based on the cumulative gradient. However, some parts of the human body are hard to see in the training data (e.g., armpits), which makes it difficult for Gaussians in these parts to be trained, and the number of Gaussians is hard to increase. Therefore, Gaussians in these parts will be relatively sparse, and we will see the hollows of the human body under novel poses (See Figure 3 in the main paper). Since we use mesh as a proxy, and Gaussians are connected to the vertices of the mesh through KNN. This makes it possible for us to explicitly increase the number of Gaussians on the body. Algorithm~\ref{algo:densification} shows our densification method. The core idea is to estimate the Gaussian density of each triangle on the mesh, and densify Gaussians according to the Gaussian density. In this way, the number of Gaussians will increase when the Gaussians are sparse in some body parts, avoiding hollow artifacts under novel poses.

Specifically, based on the KNN results $\mathcal{S}_g$, we can estimate the Gaussian density of each triangle. We set a density threshold $d_0$, and when the triangle density $d_f[i]$ is below $d_0$, the sampling probability of the triangle is set to $P[i] = \max(0, d_0-d_f[i])$. We need a total of $\mathrm{sum}(P)$ Gaussians to make the density of each triangle higher than $d_0$, but we only add a small percent $p_0$ each time we execute this algorithm. Finally, we apply multinomial sampling according to the probability $P$ to get how many Gaussians need to be added on each triangle, and randomly place Gaussians. For the density threshold and densification percent, we set $d_0=1$ and $p_0=0.02$ in our algorithm.

\begin{algorithm}[h]
\SetAlgoLined
\DontPrintSemicolon
\SetKw{Continue}{continue}
\SetKw{Break}{break}
\newcommand\mycommfont[1]{\footnotesize\sffamily\textcolor{blue}{#1}}
\SetCommentSty{mycommfont}
\KwIn{KNN index sets $\{ \mathcal{S}_g^i \}_{i \in [1,N_g]}$ \newline
    Mesh triangles $ \{f^i\}_{i \in [1,N_f]} $
    }
\KwOut{Added Gaussian positions $\{\mathbf{x}_g^i\}_{i \in [1,N_g']}$
    }
\;

$D_v \in \mathbb{Z}^{N_v}$, the number of Gaussians close to each vertex \;
$d_f \in \mathbb{R}^{N_f}$, Gaussian density for each triangle \;
$n_g \in \mathbb{Z}^{N_f}$, added Gaussian number for each triangle\;

\;
initialize $D_v, d_f$ with zero\;
\tcp{Convert KNN results $\mathcal{S}_g$ to triangle's Gaussian density $d_f$}
\For{Gaussian index $i\gets1$ \KwTo $N_g$}{
    \For{Vertex index $k \in \mathcal{S}_g^i$}{
        $D_v[k] \gets D_v[k] + 1$ \;
    }
}
\For{Triangle index $i\gets1$ \KwTo $N_f$}{
    \For{Vertex index $k \in f^i $}{
        $d_f[i] \gets d_f[i] + D_v[k]$ \;
    }
    \tcp{We assume each vertex is connected with 6 triangles}
    $d_f[i] \gets d_f[i] / (6 \times K_g)$ \;
}

\;
\tcp{Add Gaussians based on triangle's Gaussian density}
$P \in \mathbb{R}^{N_f}, P = \max(0, d_0 - d_f)$\;
\tcp{We only add a small proportion of Gaussians each time when applying this algorithm}
$N'_g \gets \mathrm{sum}(P) \times p_0$\;
$n_g \gets \mathrm{MultinomialSample}(P/\mathrm{sum}(P), N'_g)$ \;
\For{Triangle index $i\gets1$ \KwTo $N_f$}{
    Randomly add $n_g[i]$ Gaussians on the $i$th triangle\;
}

\caption{Gaussian densification based on density}
\label{algo:densification}
\end{algorithm}

\begin{figure*}[htbp]
  \begin{center}
  \includegraphics[width=0.9\textwidth]{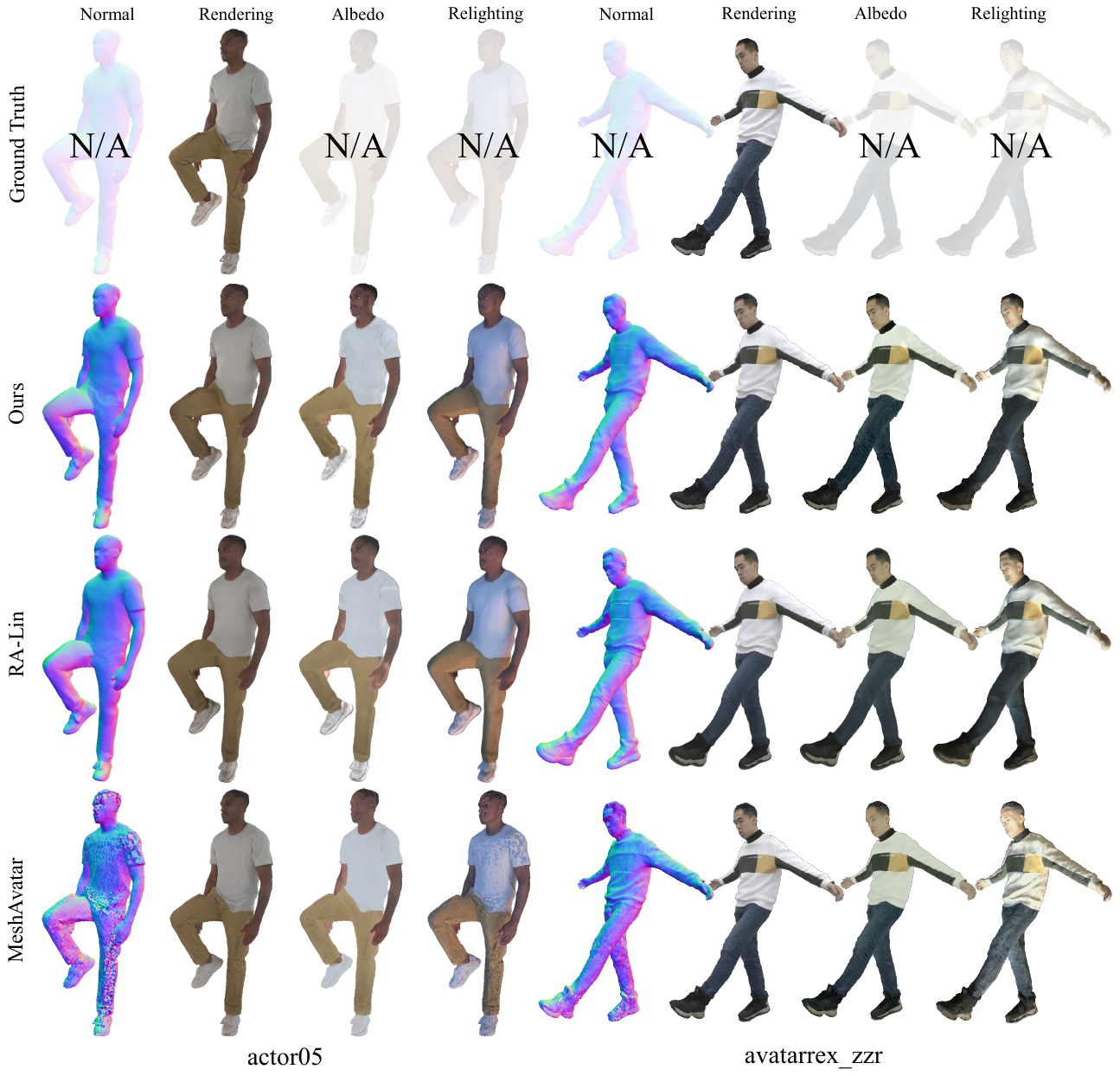}
  \end{center}
  \vspace{-1mm}
  \caption{ Qualitative comparison with RA-Lin~\cite{lin2024relightable} and MeshAvatar~\cite{chen2024meshavatar} on ActorsHQ and AvatarRex dataset.}
  \label{fig:re-compare}
\end{figure*}

\begin{figure*}[htbp]
  \centering
  \includegraphics[width=0.9\textwidth]{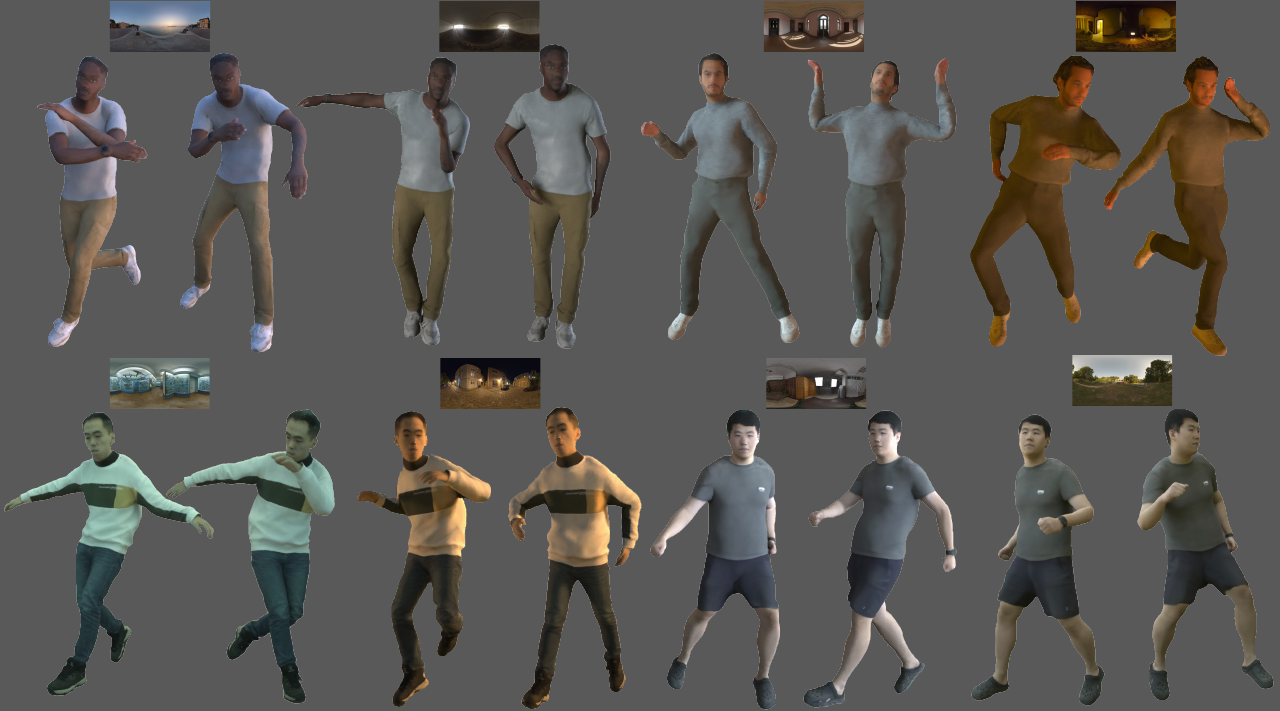}
  \vspace{-1mm}
  \caption{ Additional relighting results of our method on ActorsHQ and AvatarRex dataset.}
  \label{fig:re-gallery}
\end{figure*}

\section{Comparison and Results on Other Datasets}

We further use ActorsHQ (dataset of HumanRF~\cite{isik2023humanrf}) and
AvatarRex~\cite{zheng2023avatarrex} dataset in our experiments to validate our method on higher resolution datasets. ActorsHQ is a high-quality dataset. We select two sequences (\textsf{actor05}, \textsf{actor07}) from ActorsHQ. We
use 7 viewpoints and 150 frames for each sequence, with each image approximately 1K resolution.
AvatarRex dataset contains full-body multi-view videos. We select two sequences (\textsf{avatarrex\_zzr}, \textsf{avatarrex\_zxc}) from AvatarRex. We use 6 viewpoints and 120 frames per sequence in our experiments, with each frame approximately 1K resolution.

As these two datasets do not have groundtruth for relighting, we compare our method qualitatively with MeshAvatar~\cite{chen2024meshavatar} and RA-Lin~\cite{lin2024relightable} using the above datasets. Figure~\ref{fig:re-compare} shows the visual
results. MeshAvatar fails to recover smooth body geometry under sparse viewpoints, resulting in
noisy relighting outputs. Our relighting results are comparable to those from RA-Lin~\cite{lin2024relightable}, and outperform MeshAvatar. We also note that the rendering speed for an image of 1K resolution is very
slow for RA-Lin (RA-Lin 134.72s vs. ours 0.14s). This is because RA-Lin uses the part-wise MLPs
to estimate the visibility, which takes a lot of time.

We also present additional relighting results of our method on ActorsHQ and AvatarRex datasets
in Figure~\ref{fig:re-gallery}. All avatars are driven by novel poses. The results demonstrate that our method can
reconstruct human avatars from high-resolution datasets like ActorsHQ and AvatarRex, and produce high-quality relighting results under challenging poses. 

\begin{table*}[h]
\caption{Quantitative results of different strategies.}
\vspace{-3mm}
\footnotesize 
\begin{tabularx}{\textwidth}{l|XXXXXXXXX}
\toprule
\multirow{2}{*}{Method} & \multicolumn{3}{c}{Albedo} & \multicolumn{3}{c}{Relighting (Training poses)} & \multicolumn{3}{c}{Relighting (Novel poses)} \\
                          & PSNR $\uparrow$    & SSIM $\uparrow$   & LPIPS $\downarrow$  & PSNR $\uparrow$    & SSIM $\uparrow$   & LPIPS $\downarrow$  & PSNR $\uparrow$    & SSIM $\uparrow$   & LPIPS $\downarrow$  \\ \hline
Ours                      & {34.3398} & {0.9580} & {0.1985} & {36.1024} & {0.8826} & {0.1647} & 28.3886 & {0.8546} & {0.1815} \\
Predict displacement, scale, rotation    & 34.4069 & 0.9581 &  0.1984  & 36.1664 & 0.8827 & 0.1652 & 28.3171  & 0.8543 &  0.1817 \\ 
Optimize opacity    & 34.2526 & 0.9576 & 0.1983  & 36.0631 & 0.8825 & 0.1647 & 28.3726 & 0.8546 & 0.1816 \\ 
\bottomrule
\end{tabularx}
\label{table:predsr}
\end{table*}

\section{Additional Ablation Study}

\begin{figure}[t]
  \begin{center}
    \includegraphics[width=0.3\textwidth]{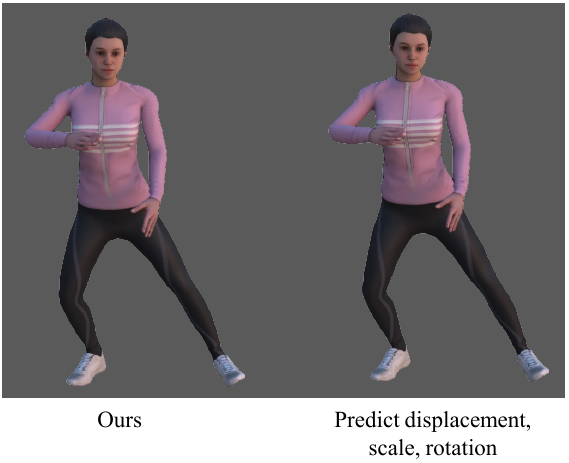}
  \end{center}
  \vspace{-5mm}
  \caption{ Qualitative results of predicting displacements, Gaussian rotation offsets and scale offsets.}
  \label{fig:re-predsr}
\end{figure}

\textbf{Predicting displacements, Gaussian rotation offsets and scale offsets.} We only predict the displacements in our method. We further follow 4D Gaussian splatting~\cite{wu20244d} to predict displacements, Gaussian rotation offsets and scale offsets to validate if this design can improve the results. Table~\ref{table:predsr} illustrates the results. The quantitative results are comparable to those without predicting the changes of rotation and scaling. We also present a qualitative result, shown in Figure~\ref{fig:re-predsr}. Both images are similar. So we choose to only predict the displacements to reduce the computational burden.

\begin{figure}[h]
  \begin{center}
    \includegraphics[width=0.45\textwidth]{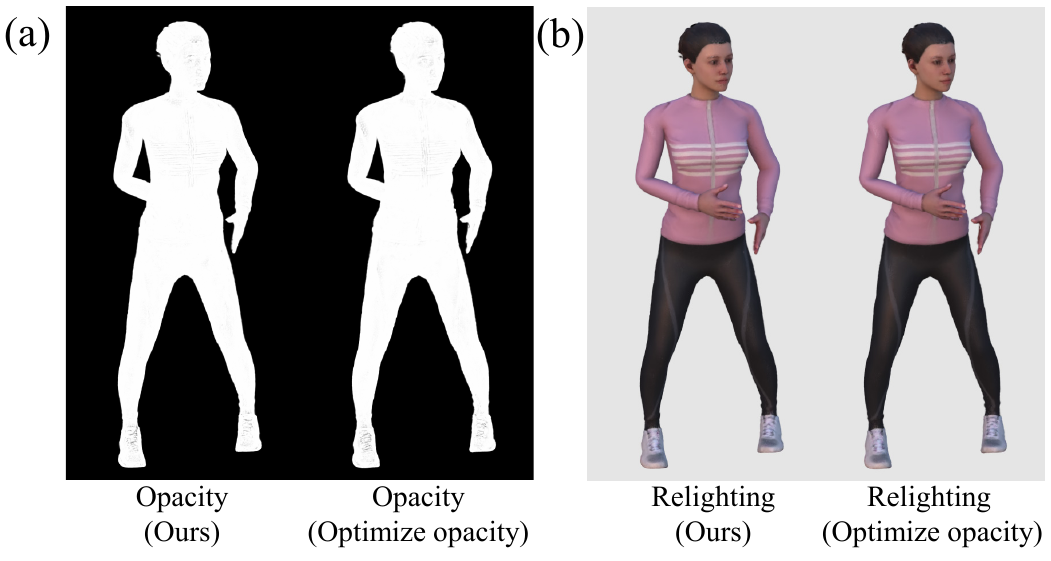}
  \end{center}
  \vspace{-5mm}
  \caption{  Qualitative results of optimizing the opacity.}
  \label{fig:re-opacity}
\end{figure}

\textbf{Optimizing opacity.} We always set opacity to 1 in our method. We further set the opacity as a learnable parameter and see if this will result in translucent artifacts. Figure~\ref{fig:re-opacity} (a) shows the opacity map of two designs, and Figure~\ref{fig:re-opacity} (b) presents the relighting results. As shown in the figure, learnable opacity does not result in translucent artifacts, and the relighting results of optimizing the opacity are almost the same as those from the proposed method. We also present quantitative results of optimizing the opacity in Table~\ref{table:predsr}. The quantitative results are also comparable to those without optimizing the opacity. Since optimizing the opacity does not significantly improve the results, we choose to set the opacity to 1. 

\end{document}